\documentclass{article}

\PassOptionsToPackage{numbers, compress}{natbib}

\usepackage[final]{neurips_2025}

\usepackage[utf8]{inputenc} 
\usepackage[T1]{fontenc}    
\usepackage{hyperref}       
\usepackage{url}            
\usepackage{booktabs}       
\usepackage{amsfonts}       
\usepackage{nicefrac}       
\usepackage{microtype}      
\usepackage{xcolor}         

\usepackage{wrapfig}
\usepackage{graphicx}
\usepackage{subfigure}
\usepackage{enumitem}
\usepackage{amsmath}
\usepackage{colortbl}
\usepackage{soul}
\usepackage{adjustbox}
\usepackage{algorithm}
\usepackage{algorithmic}
\usepackage{caption}
\usepackage{makecell}
\usepackage{multirow}
\usepackage{amsthm}
\usepackage{array}
\usepackage{longtable}

\usepackage[toc,page,header]{appendix}
\usepackage{minitoc}

\definecolor{DarkGreen}{rgb}{0, 0.8039, 0.4}
\definecolor{LightGreen}{rgb}{0.9450980392, 0.768627451, 0.0588235294115}
\definecolor{Red}{rgb}{0.85098, 0.0117647, 0.4078431373}

\definecolor{color_links}{rgb}{0, 0.5019607843, 0.2509803922}
\hypersetup{
    colorlinks=true,
    linkcolor=color_links,      
    citecolor=color_links,   
    urlcolor=color_links   
}

\title{M3DMap: Object-aware Multimodal 3D Mapping for Dynamic Environments}

\author{%
  D.A. Yudin\\
  Moscow Institute of Physics and Technology, Moscow, Russia\\
  AIRI, Moscow, Russia \\
  \texttt{yudin.da@mipt.ru} \\
}

\begin{document}

\doparttoc
\faketableofcontents


\maketitle

\begin{abstract}
3D mapping in dynamic environments poses a challenge for modern researchers in robotics and autonomous transportation. There are no universal representations for dynamic 3D scenes that incorporate multimodal data such as images, point clouds, and text. This article takes a step toward solving this problem.  
It proposes a taxonomy of methods for constructing multimodal 3D maps, classifying contemporary approaches based on scene types and representations, learning methods, and practical applications. Using this taxonomy, a brief structured analysis of recent methods is provided.  
The article also describes an original modular method called M3DMap, designed for object-aware construction of multimodal 3D maps for both static and dynamic scenes. It consists of several interconnected components: a neural multimodal object segmentation and tracking module; an odometry estimation module, including trainable algorithms; a module for 3D map construction and updating with various implementations depending on the desired scene representation; and a multimodal data retrieval module. The article highlights original implementations of these modules and their advantages in solving various practical tasks, from 3D object grounding to mobile manipulation.  
Additionally, it presents theoretical propositions demonstrating the positive effect of using multimodal data and modern foundational models in 3D mapping methods.  
Details of the taxonomy and method implementation are available at \url{https://yuddim.github.io/M3DMap}.

\end{abstract}

\section{Introduction}\label{sec:intro}
In recent years, various foundational neural models for image recognition (e.g., CLIP \cite{radford2021clip}, LSeg~\cite{li2022lseg}, SAM \cite{kirillov2023SAM}, DINO \cite{oquab2023dinov2}) and 3D point clouds (Uni3D \cite{zhou2023uni3d}, Sonata \cite{wu2025sonata}, etc.) have rapidly advanced. This has led to the emergence and development of approaches for enriching 3D maps with additional information, such as feature representations of objects aligned with textual modalities, enabling the inclusion of arbitrary objects beyond a fixed set of categories.  

However, there is no unified approach to representing 3D scenes in such multimodal maps: some works use point clouds/voxels \cite{jatavallabhula2023conceptfusion, huang2023avlmaps}, bird's-eye-view projections \cite{zhang2022beverse, qi2025gpt4scene}, while others employ Neural Radiance Fields (NeRF) \cite{kerr2023lerf}, Signed Distance Fields (SDF) \cite{yamazaki2024openfusion}, Gaussian splatting \cite{qin2024langsplat}. A significant body of research is based on representing scenes as various types of graphs \cite{gu2024conceptgraphs, werby2024hierarchicalov3d, zhang2025openfun, wu2025usg}. This diversity is largely due to the wide range of applications, from navigation and control of intelligent robots and autonomous vehicles to 3D/4D scene reconstruction in Virtual/Augmented Reality systems.  

Moreover, there is a noticeable bias toward methods based on image or video modalities, with significantly less attention paid to point cloud data, despite its growing importance due to the decreasing cost of LiDARs. There are no unified datasets or benchmarks that simultaneously address tasks such as recognition (segmentation), tracking, object description in 3D space, question-answering about dynamic 3D scenes, or using maps for robot planning and control. However, multimodal datasets based on established benchmarks like ScanNet \cite{dai2017scannet} and 3DScan \cite{wald20193rscan} are gradually emerging.  

This work attempts to systematize existing research in multimodal 3D mapping, covering all major scene representations and key applications. Special attention is given to dynamic scenes, where there is a notable lack of research, benchmarks, and technical solutions.  

Additionally, based on extensive research experience in this field, a methodology is proposed that integrates various computer vision methods into a unified framework for multimodal 3D mapping, adaptable to diverse practical tasks using different data modalities and scene representations.  

\textbf{Contributions of the article can be summarized as follows:}  
\begin{itemize}  
    \item A categorization of approaches for constructing multimodal 3D maps, including those for dynamic environments, is proposed. It is detailed in Section \ref{sec:related_works}.  
    \item A modular approach called M3DMap is introduced for building multimodal 3D maps. It uses neural models for object extraction and tracking, trainable odometry estimation, algorithms for constructing accumulated and instantaneous map representations, and methods for retrieving multimodal data (see Section \ref{sec:proposed_approach}). Experiments show that the proposed modules significantly improve performance in real-world downstream tasks in robotics and autonomous transportation (see Section \ref{sec:effect}).  
    \item Theoretical propositions are presented, demonstrating the improvement in object recognition quality when constructing 3D maps by leveraging multiple modalities or additional encoders for raw sensor data (see Section \ref{sec:theory}).  
\end{itemize}  

\section{Categorization of Related Approaches}\label{sec:related_works}  

\begin{figure}[t]  
  \centering  
  \includegraphics[width=0.99\textwidth]{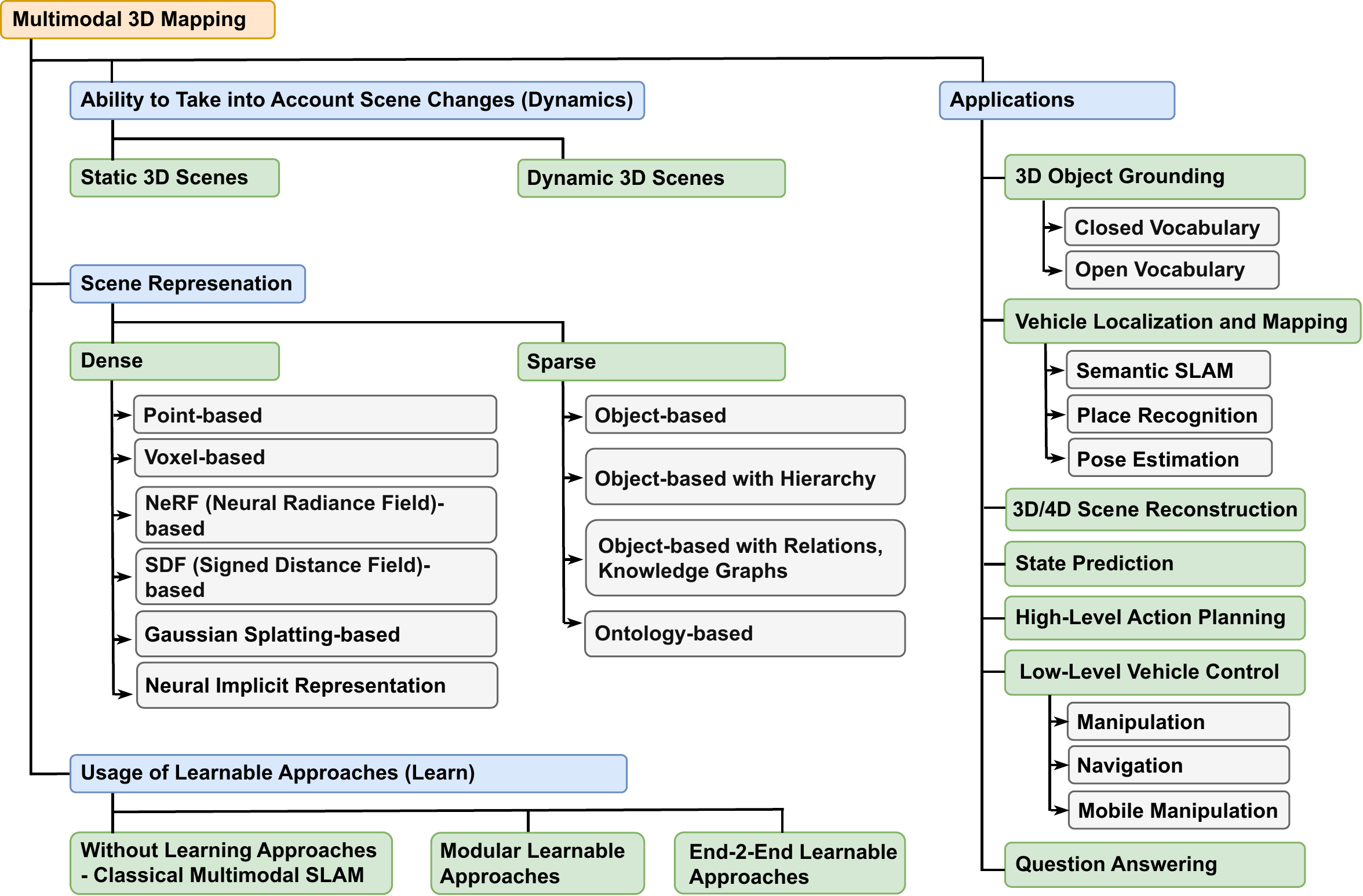}  
  \caption{Proposed taxonomy of methods for constructing multimodal 3D maps}  
  \label{fig:taxonomy}  
\end{figure}

This article proposes a taxonomy for systematizing existing research in multimodal 3D mapping, as shown in Figure \ref{fig:taxonomy}.  

The taxonomy categorizes methods based on four main criteria:  
\begin{itemize}  
    \item \textbf{Ability to Take into Account Scene Changes}: Two broad categories are considered—static scenes (where all objects are stationary) and dynamic 3D scenes (including three types of dynamics: objects moving within the robot's field of view, outside it, or interactive objects that can open/close, turn on/off, etc.).  
    \item \textbf{Scene Representation}: This criterion extends the classification proposed in \cite{yudin2024multimodalmapping} with Neural Implicit Scene Representation, where the map is implicitly formed as an updatable tensor from raw input data. Such representations are typically used for planning and control tasks in robotics.  
    \item \textbf{Usage of Learnable Approaches}: This criterion also considers methods that do not use deep learning, often aimed at solving practical robot localization tasks. Distinguishing between modular and end-to-end approaches can be challenging; here, end-to-end methods are defined as single neural models where gradients can flow from output to raw input data during training.  
    \item \textbf{Applications}: Seven key practical domains are highlighted. Tasks like 3D Object Grounding include detection and segmentation in 3D scenes, with open-vocabulary approaches dominating as they accommodate natural language queries. Applications related to SLAM, pose estimation, and place recognition are grouped under Vehicle Localization and Mapping. 4D scene reconstruction adds a fourth dimension—time—to spatial coordinates. State Prediction includes approaches that generate maps or estimate the robot's future state. High-Level Action Planning involves using multimodal 3D map representations with Large Language Models (LLMs) to generate action plans. Low-Level Vehicle Control generates specific robot actions for manipulation or navigation tasks. Question-Answering applications encompass a wide range of subtasks and question types about 3D scenes, including captioning and reasoning.  
\end{itemize}  

A brief analysis of recent works, particularly those from 2023–2025 with open-source code, is provided below. The primary categorization criterion is the handling of scene dynamics.  

\textbf{Dynamic 3D Scenes.} Methods in this category are systematized in Tables \ref{tab:dense_dynamic} and \ref{tab:sparse_dynamic}.  

\begin{table}[!h]
    \caption{Modern Dense Methods for Constructing Multimodal 3D Maps of Dynamic Scenes. The prefix "D:" in scene representation categories stands for Dense}
    \label{tab:dense_dynamic}
    \centering
    \tiny
    \begin{tabular}{p{2.3cm}|p{0.5cm}|p{2.5cm}|p{1.5cm}|p{2.5cm}|p{0.5cm}|p{2.5cm}}
        \hline 
        \rule{0mm}{4mm}Method & Year & Scene Representation & Learn & Application & Code & Map Modalities \\  
        \hline 
        \hline 
        \rule{0mm}{4mm}DynamicGSG \cite{ge2025dynamicgsg} & 2025 & D: Gaussian Splatting-based & Modular & Object Grounding, 3D Scene Reconstruction & \checkmark & Gaussians, Image-Text features, Hierarchical 3D Scene Graph \\  \hline 
        \rule{0mm}{4mm}4D-Gaussian-Splatting \cite{yang20244dgs} & 2024 & D: Gaussian Splatting-based & End-2-End & 4D Scene Reconstruction & \checkmark & Color, gaussian features, time \\ \hline 
        \rule{0mm}{4mm}4D-GS \cite{wu20244d} & 2024 & D: Gaussian Splatting-based & End-2-End & 4D Scene Reconstruction & \checkmark & Color, gaussian features, time \\ \hline  
        \rule{0mm}{4mm}GeFF \cite{qiu2024geff} & 2024 & D: NeRF-based & End-2-End & 3D Scene Reconstruction, Mobile Manipulation & \checkmark & Color, Density, Visual-Language Features \\ \hline 
        \rule{0mm}{4mm}Lifelong LERF \cite{rashid2024lifelonglerf} & 2024 & D: NeRF-based & End-2-End & Object Grounding, 3D Scene Reconstruction & ~ & CLIP Features, Color \\ \hline 
        \rule{0mm}{4mm}Remembr \cite{anwar2024remembr} & 2025 & D: Neural Implicit Representation & End-2-End & Object Grounding, High-Level Action Planning, Question Answering & \checkmark & Text-image features, text, position, time \\ \hline 
        \rule{0mm}{4mm}OmniDrive \cite{wang2025omnidrive} & 2025 & D: Neural Implicit Representation & End-2-End & Path Planning, Question Answering & \checkmark & 2D VLM LLAVA features,  3D BEV features, 3D Pose Encoding \\ \hline 
        \rule{0mm}{4mm}BEVDriver \cite{winter2025bevdriver} & 2025 & D: Neural Implicit Representation & End-2-End & Object Grounding, High-Level Action Planner, Path Planning, Vehicle Navigation & ~ & BEVMap of LiDAR PointCloud features, Multi-view Image Features, Navigation Text Instructions \\ \hline 
        \rule{0mm}{4mm}DriveTransformer \cite{jia2025drivetransformer} & 2025 & D: Neural Implicit Representation & End-2-End & Object Grounding, State Prediction, Path Planning & \checkmark & Pose features, Multi-view image features, Memory features \\ \hline 
        \rule{0mm}{4mm}Embodied-Reasoner \cite{zhang2025embodiedreasoner} & 2025 & D: Neural Implicit Representation & End-2-End & Object Grounding, High-Level Action Planning for Mobile Manipulation, Question Answering & \checkmark & Image features, Text features, Action features, Thought features \\ \hline 
        \rule{0mm}{4mm}NPField \cite{alhaddad2024npfield} & 2024 & D: Neural Implicit Representation & End-2-End & Vehicle Navigation & \checkmark & Pose, Occupancy, Footprint features \\ \hline 
        \rule{0mm}{4mm}UniAD \cite{hu2023uniad} & 2023 & D: Neural Implicit Representation & End-2-End & Object Grounding, State Prediction, Path Planning & \checkmark & Multi-view image, HD-Map , Object, Motion, Occupancy features \\ \hline 
        \rule{0mm}{4mm}VAD \cite{jiang2023vad} & 2023 & D: Neural Implicit Representation & End-2-End & Path Planning, State Prediction & \checkmark & HD-Map Features, Muti-view image features \\ \hline 
        \rule{0mm}{4mm}FAST3R \cite{yang2025fast3r} & 2025 & D: Point-based & End-2-End & 4D/3D Scene Reconstruction, Vehicle Localization & \checkmark & Image, Depth, Pose features \\ \hline 
        \rule{0mm}{4mm}LMR in Prior 3D LiDAR Maps \cite{belkin2024lmrinprior} & 2024 & D: Point-based & Modular & Vehicle Localization & ~ & Depth, Pose, Visual Point Cloud \\ \hline 
        \rule{0mm}{4mm}Mask3D-FM (SceneFun3D) \cite{delitzas2024scenefun3d} & 2024 & D: Point-based & Modular & Object Grounding, State Prediction & \checkmark & Point cloud, color, Image-text features, Motion vector, Object labels \\ \hline 
        \rule{0mm}{4mm}WildFusion \cite{liu2024wildfusion} & 2024 & D: SDF-based & End-2-End & Object Grounding, Path Planning & \checkmark & Pixel-level geometry, color, semantics , and traversability \\ \hline 
        \rule{0mm}{4mm}BEVerse \cite{zhang2022beverse} & 2022 & D: Voxel-based & End-2-End & Object Grounding, State Prediction, Vehicle Outdoor Navigation & \checkmark & Image features, BEV features, Object label features, HD-Map features \\ \hline
    \end{tabular}
\end{table}

\begin{table}[!h]
    \caption{Modern Sparse Methods for Constructing Multimodal 3D Maps of Dynamic Scenes. The prefix "S:" in scene representation categories stands for Sparse}
    \label{tab:sparse_dynamic}
    \centering
    \tiny
    \begin{tabular}{p{2.3cm}|p{0.5cm}|p{2.2cm}|p{1.5cm}|p{2.5cm}|p{0.5cm}|p{3.5cm}}
    \hline
        \rule{0mm}{4mm}Method & Year & Scene Representation & Learn & Application & Code & Map Modalities \\ \hline
        \hline
        \rule{0mm}{4mm}OpenFunGraph \cite{zhang2025openfun} & 2025 & S: Knowledge graphs & Modular & Object Grounding & \checkmark & Interactive Functional Scene Graph \\ \hline
        \rule{0mm}{4mm}Motion-aware Contrastive Learning \cite{nguyen2025motion} & 2025 & S: Knowledge graphs & End-2-End & Object Grounding & ~ & Images, Labels, Edges, Node features \\ \hline
        \rule{0mm}{4mm}Gemini Robotics \cite{team2025gemini} & 2025 & S: Knowledge graphs & End-2-End & Question Answering, Object Grounding, Action Planning for Mobile Manipulation & \checkmark & Imates, Text, 3D-pose \\ \hline
        \rule{0mm}{4mm}MM2SG (MM-OR) \cite{ozsoy2025MM2SG} & 2025 & S: Knowledge graphs & End-2-End & Object Grounding, State Prediction & \checkmark & Features of RGB-D, PointCloud, Audio, Tracking, Memory Scene Graphs, Segmentation Masks \\ \hline
        \rule{0mm}{4mm}USG \cite{wu2025usg} & 2025 & S: Knowledge graphs & Modular & Object Grounding & \checkmark & Image, Pointcloud, Text, Object sequence, 3D Scene Graph \\ \hline
        \rule{0mm}{4mm}AriGraph \cite{anokhin2024arigraph} & 2024 & S: Knowledge graphs & Modular & Action Planning, Question Answering & \checkmark & Text Graph with Updated nodes and edges \\ \hline
        \rule{0mm}{4mm}3d vsg \cite{looper20233dvsg} & 2023 & S: Knowledge graphs & Modular & State Prediction & \checkmark & Graph Object, Interaction Features, State/Position/Instance Variability \\ \hline
        \rule{0mm}{4mm}NEP \cite{kurenkov2023nep} & 2023 & S: Knowledge graphs & Modular & Object Grounding & \checkmark & Scene graph: node and edge features \\ \hline
        \rule{0mm}{4mm}Dyted \cite{zhang2023dyted} & 2023 & S: Knowledge graphs & Modular & State Prediction & \checkmark & discrete-time dynamic graph \\ \hline
        \rule{0mm}{4mm}Labrad-OR \cite{ozsoy2023labrad} & 2023 & S: Knowledge graphs & End-2-End & State Prediction & \checkmark & Pontcloud, image features, scene graph \\ \hline
        \rule{0mm}{4mm}PSG4D \cite{yang20234dpsg} & 2023 & S: Knowledge graphs & Modular & Object Grounding, High-Level Action Planning, Question Answering & \checkmark & Point cloud, 3D Scene Graph (Class Labels, IDs Relations, Time) \\ \hline
        \rule{0mm}{4mm}MCGNet (RealGraph) \cite{lin2023realgraph} & 2023 & S: Knowledge graphs & End-2-End & Object Grounding, State Prediction & \checkmark & Image features, 3D boxes, Context Graph \\ \hline
        \rule{0mm}{4mm}Spatio-Temporal Object Tracking  \cite{patel2022proactive} & 2022 & S: Knowledge graphs & Modular & State Prediction & \checkmark & Dynamic Scene graph: node and edge features \\ \hline
        \rule{0mm}{4mm}3D-Mem \cite{yang20253dmem} & 2025 & S: Object-based & Modular & Question Answering, Object Grounding & \checkmark & Images, Text Descriptions,  \\ \hline
        \rule{0mm}{4mm}SUGAR \cite{chen2024sugar} & 2024 & S: Object-based & End-2-End & Object Grounding, Manipulation & \checkmark & Pointcloud, Text features, Pose, Instance Mask \\ \hline
        \rule{0mm}{4mm}UP-VL \cite{najibi2023unsupervised3dperception} & 2023 & S: Object-based & End-2-End & Object Grounding & ~ & Pointcloud, Text features, Image Features, Object Pose \\ \hline
        \rule{0mm}{4mm}D3A \cite{idreesbuilding2023D3A} & 2023 & S: Object-based & Modular & Object Grounding & ~ & Images, Features, Locations \\ \hline
        \rule{0mm}{4mm}Moma-LLM \cite{honerkamp2024languagegroundeddynamicscene} & 2024 & S: Object-based with Hierarchy & Modular & Object Grounding, High-Level Action Planning for Mobile Manipulation & \checkmark & Hierarchical Graph, Semantic labels, Image features, Localization data \\ \hline
        \rule{0mm}{4mm}Khronos \cite{schmid2024khronos} & 2024 & S: Object-based with Hierarchy & Modular & Object Grounding, Vehicle Localization & \checkmark & Image, Depth, Semantic Mask, Features, Locations, Poses \\ \hline
        \rule{0mm}{4mm}GraphEQA \cite{saxena2024grapheqa} & 2024 & S: Object-based with Hierarchy & Modular & Question Answering, Object Grounding, High-Level Action Planner & \checkmark & Image, Hierarchical Graph, Semantic Masks, Pose, History Answers \\ \hline
        \rule{0mm}{4mm}LP2 \cite{gorlo2024lp2} & 2024 & S: Object-based with Hierarchy & Modular & State Prediction & \checkmark & Hierarchical 3D Scene Graph \\ \hline
        \rule{0mm}{4mm}DSG (Kimera) \cite{rosinol2020dsg} & 2020 & S: Object-based with Hierarchy & Modular & Object Grounding, Vehicle Localization & \checkmark & Metric-Semantic Mesh, Places, Object Labels, Human tracks, Room and Building Labels \\ \hline
    \end{tabular}
\end{table}
Many approaches represent video sequences as sequences of scene graphs for each frame. Such methods are not considered here. Instead, the focus is on methods that explicitly account for or generate 3D information about detected objects.  

Table \ref{tab:dense_dynamic} groups contemporary dense methods for constructing multimodal 3D maps of dynamic scenes. It shows that end-to-end learning dominates these approaches, largely due to higher final task performance but also demanding significant computational resources for training.  

Among dense methods for 4D and 3D scene reconstruction, Gaussian splatting (4D-GS \cite{wu20244d}, 4D-Gaussian-Splatting \cite{yang20244dgs}, DynamicGSG \cite{ge2025dynamicgsg}) and NeRF (Lifelong LERF \cite{rashid2024lifelonglerf}, GeFF \cite{qiu2024geff}) are prominent. GeFF \cite{qiu2024geff} also uses NeRF scene representations for mobile manipulation tasks.  

Neural Implicit Representation methods primarily build implicit memory storing embeddings that encode 3D scenes for planning and robot navigation (Remembr \cite{anwar2024remembr}, Embodied-Reasoner \cite{zhang2025embodiedreasoner}, NPField \cite{alhaddad2024npfield}) or autonomous vehicles (OmniDrive \cite{wang2025omnidrive}, BEVDriver \cite{winter2025bevdriver}, DriveTransformer \cite{jia2025drivetransformer}, UniAD \cite{hu2023uniad}, VAD \cite{jiang2023vad}).  

Dense point-based methods (LMR in Prior 3D LiDAR Maps \cite{belkin2024lmrinprior}, Mask3D-FM \cite{delitzas2024scenefun3d}) are mainly applied in dynamic scenes for vehicle localization.  

SDF-based (WildFusion \cite{liu2024wildfusion}) and voxel-based (BEVerse \cite{zhang2022beverse}) methods are developed for path planning and vehicle state prediction. Notably, Bird's-Eye-View (BEV) representations, as in BEVerse \cite{zhang2022beverse}, are a de facto standard in modern autonomous vehicle navigation models.  

For 3D Object Grounding, approaches based on all dense scene representations exist. However, most benchmarks and research focus on outdoor scenes, while indoor robotics still lacks annotated dynamic scene datasets.  

Table \ref{tab:sparse_dynamic} lists modern sparse methods for constructing multimodal 3D maps of dynamic scenes. Their number slightly exceeds dense methods, reflecting the rapid development of 3D scene graph representations. Object-based approaches are also graph-based but omit edges representing relationships between objects.  

Knowledge graphs and object-based representations enable the effective use of Large Language Models (LLMs) and Vision-Language Models (VLMs) for tasks like question answering, as seen in Gemini Robotics \cite{team2025gemini}, AriGraph \cite{anokhin2024arigraph}, 3D-Mem \cite{yang20253dmem}, Moma-LLM \cite{honerkamp2024languagegroundeddynamicscene}, GraphEQA \cite{saxena2024grapheqa}, and PSG4D \cite{yang20234dpsg}.  

Several methods predict future states of 3D scene graphs, such as MM2SG \cite{ozsoy2025MM2SG}, Labrad-OR \cite{ozsoy2023labrad}, 3d vsg \cite{looper20233dvsg}, Dyted \cite{zhang2023dyted}, and LP2 \cite{gorlo2024lp2}.  

Object grounding and classification tasks are addressed by nearly all methods in this group. Few specialize in robotic applications, such as Khronos \cite{schmid2024khronos} and DSG \cite{rosinol2020dsg} for robot localization, SUGAR \cite{chen2024sugar} for manipulation, and Gemini Robotics \cite{team2025gemini} and Moma-LLM \cite{honerkamp2024languagegroundeddynamicscene} for mobile manipulation.  

Existing sparse methods for dynamic scenes still suffer from limited annotated datasets but hold great promise due to their compact representation of data sequences and symbolic nature, aligning well with LLMs, which underpin most AI systems.  

\textbf{Static 3D Scenes.} Methods in this category are systematized in Tables \ref{tab:dense_static} and \ref{tab:sparse_static}. Notably, there are far more annotated datasets for static scenes than dynamic ones, leading to greater diversity in methods.  

Table \ref{tab:dense_static} lists recent dense methods for constructing multimodal 3D maps of static scenes. Most pre-2024 works are analyzed in \cite{yudin2024multimodalmapping}.  

\begingroup\tiny\begin{longtable}[t]{p{2.3cm}|p{0.5cm}|p{2.2cm}|p{1.5cm}|p{2.5cm}|p{0.5cm}|p{3.5cm}}
    \caption{Modern Dense Methods for Constructing Multimodal 3D Maps of Static Scenes\label{tab:dense_static}}\\ 
    \hline
    \rule{0mm}{4mm}Method & Year & Scene Representation & Learn & Application & Code & Map Modalities \\ \hline
    \endfirsthead		
    \caption[]{(Continuation)} \\
    \hline
    \rule{0mm}{4mm}Method & Year & Scene Representation & Learn & Application & Code & Map Modalities \\ \hline
    \endhead
    \rule{0mm}{4mm}M3 \cite{zou2025m3} & 2025 & D: Gaussian Splatting-based & End-2-End & 3D Scene Reconstruction, Object Grounding & \checkmark & RGB and foundation model (CLIP/SigLIP, LLAMA3/V, DINOv2, SEEM) embeddings \\ \hline
    \rule{0mm}{4mm}SplatTalk \cite{thai2025splattalk} & 2025 & D: Gaussian Splatting-based & End-2-End & Question Answering  &  & Gaussians, Image-Text features \\ \hline
    \rule{0mm}{4mm}GaussianGraph \cite{wang2025gaussiangraph} & 2025 & D: Gaussian Splatting-based & Modular & Object Grounding & \checkmark & Gaussian Point Cloud, Object clusters, CLIP features, Attributes, Relations, Distances \\ \hline
    \rule{0mm}{4mm}OpenGaussian \cite{wu2024opengaussian} & 2024 & D: Gaussian Splatting-based & End-2-End & 3D Scene Reconstruction, Object Grounding & \checkmark & 3D Gaussians, 3D codebook-based features, CLIP-features \\ \hline
    \rule{0mm}{4mm}LangSplat \cite{qin2024langsplat} & 2024 & D: Gaussian Splatting-based & End-2-End & 3D Scene Reconstruction, Object Grounding & \checkmark & 3D Gaussians, Encoded CLIP features \\ \hline
    \rule{0mm}{4mm}Semantic Gaussians \cite{guo2024semanticgaussians} & 2024 & D: Gaussian Splatting-based & Modular & 3D Scene Reconstruction, Object Grounding & \checkmark & CLIP, Lseg features \\ \hline
    \rule{0mm}{4mm}LLM-grounder \cite{yang2024llmgrounder} & 2024 & D: NeRF-based & Modular & Object Grounding, High-Level Action Planning, Question Answering & \checkmark & CLIP Features, Color, Pointcloud \\ \hline
    \rule{0mm}{4mm}uSF \cite{skorokhodov2024usf} & 2024 & D: NeRF-based & Modular & 3D Scene Reconstruction & \checkmark & Color, label, uncertainty \\ \hline
    \rule{0mm}{4mm}3d-llm \cite{hong20233dllm} & 2023 & D: NeRF-based & Modular & Object Grounding, Question Answering & \checkmark & Pointcloud, Color, EVACLIP 3D Features, \rule{0mm}{4mm}Neural voxel field \\ \hline
    \rule{0mm}{4mm}LERF \cite{kerr2023lerf} & 2023 & D: NeRF-based & End-2-End & Object Grounding, 3D Scene Reconstruction & \checkmark & Color, Density, DINO, CLIP features \\ \hline
    \rule{0mm}{4mm}3D-OVS \cite{liu2023weaklysupervised3dseg} & 2023 & D: NeRF-based & End-2-End & Object Grounding & \checkmark & Color, Density, DINO, CLIP features \\ \hline
    \rule{0mm}{4mm}SpatialLM \cite{spatiallm2025} & 2025 & D: Neural Implicit Representation & End-2-End & Object Grounding & \checkmark & Color Pointcloud, Text features \\ \hline
    \rule{0mm}{4mm}EmbodiedScan \cite{wang2024embodiedscan} & 2024 & D: Neural Implicit Representation & End-2-End & Object Grounding, 3D Scene Reconstruction & \checkmark & Point cloud, Image, Visual-Language features \\ \hline
    \rule{0mm}{4mm}SpatialVLM \cite{chen2024spatialvlm} & 2024 & D: Neural Implicit Representation & End-2-End & Question Answering  & \checkmark & Image, depth, pointcloud, object category and caption features \\ \hline
    \rule{0mm}{4mm}LL3DA \cite{chen2024ll3da} & 2024 & D: Neural Implicit Representation & End-2-End & Object Grounding, High-Level Action Planning, Question Answering & \checkmark & Features of Color Point cloud, Text, Visual Prompts \\ \hline
    \rule{0mm}{4mm}Dynamic implicit representations \cite{marza2023dynamicimplicitrep} & 2023 & D: \rule{0mm}{4mm}Neural Implicit Representation & End-2-End & Vehicle Navigation & \checkmark & RGB-D  and segmentation mask features \\ \hline
    \rule{0mm}{4mm}robo-vln \cite{irshad2021robovln} & 2021 & D: Neural Implicit Representation & End-2-End & Robot Indoor Navigation & \checkmark & Image, Depth, Text Features \\ \hline
    \rule{0mm}{4mm}GPT4Scene \cite{qi2025gpt4scene} & 2025 & D: Point-based & Modular & Object Grounding, High-Level Action Planning, Question Answering & \checkmark & Image with object marks, BEV features, Text features \\ \hline
    \rule{0mm}{4mm}Locate-3D \cite{arnaud2025locate3d} & 2025 & D: Point-based & Modular & Object Grounding, High-Level Action Planning, Question Answering & \checkmark & CLIP, DINO Features, Color, Pointcloud \\ \hline
    \rule{0mm}{4mm}R3LIVE++ \cite{lin2024r3live} & 2024 & D: Point-based & Without learning & 3D Scene Reconstruction, Vehicle Localization & \checkmark & Point cloud, Radiance map, Color \\ \hline
    \rule{0mm}{4mm}RegionPLC \cite{yang2024regionplc} & 2024 & D: Point-based & Modular & Object Grounding, High-Level Action Planning, Question Answering & \checkmark & Pointcloud, color, text \\ \hline
    \rule{0mm}{4mm}SeCG \cite{xiao2024secg} & 2024 & D: Point-based & Modular & Object Grounding & \checkmark & Point cloud, Color, Semantic labels, Text features,  Implicit Relation Graph Features \\ \hline
    \rule{0mm}{4mm}ConceptFusion\cite{jatavallabhula2023conceptfusion} & 2023 & D: Point-based & Modular & Object Grounding & \checkmark & Pointcloud, Image-text features, Color \\ \hline
    \rule{0mm}{4mm}OpenMask3D\cite{takmaz2023openmask3d} & 2023 & D: Point-based & Modular & Object Grounding & \checkmark & Pointcloud, Image-text features, Color \\ \hline
    \rule{0mm}{4mm}Openscene\cite{peng2023openscene} & 2023 & D: Point-based & End-2-End & Object Grounding & \checkmark & Image, Pointcloud, Text Features \\ \hline
    \rule{0mm}{4mm}PLA\cite{ding2023pla} & 2023 & D: Point-based & Modular & Object Grounding & \checkmark & Pointcloud, color, text \\ \hline
    \rule{0mm}{4mm}CLIP-FO3D \cite{zhang2023clipfo3d} & 2023 & D: Point-based & Modular & Object Grounding & ~ & Point cloud, Color, CLIP features \\ \hline
    \rule{0mm}{4mm}OVIR-3D \cite{lu2023ovir} & 2023 & D: Point-based & Modular & Object Grounding, Manipulation & \checkmark & Pointcloud, Color, Image-text (Detic) features, Object labels \\ \hline
    \rule{0mm}{4mm}NLMaps\cite{chen2023nlmaps} & 2022 & D: Point-based & Modular & Object Grounding, High-Level Action Planning & \checkmark & Image, Visual-text feature, Occupancy \\ \hline
    \rule{0mm}{4mm}CLIP-Fields \cite{shafiullah2022clipfields} & 2022 & D: Point-based & Modular & Object Grounding, Indoor Vehicle Navigation, Place Recognition & \checkmark & Point cloud, Color, CLIP features, Semantic label \\ \hline
    RTabMap \cite{labbe2019rtabmap} & 2019 & D: Point-based & Without learning & Vehicle Localization & \checkmark & Pointcloud, Image, Octomap, 2D Occupancy, Pose, Map Graph \\ \hline
    \rule{0mm}{4mm}Scene-LLM \cite{fu2024scenellm} & 2024 & D: Point-Voxel-based & Modular & High-Level Action Planning, Question Answering & ~ & Pointcloud, CLIP Features, Text Captions  \\ \hline
    \rule{0mm}{4mm}UPPM \cite{mdfaa2024uppm} & 2024 & D: SDF-based & Modular & Object Grounding, 3D Scene Reconstruction & \checkmark & Color, Density, Semantic/Instance Labels \\ \hline
    \rule{0mm}{4mm}Open-Fusion \cite{yamazaki2024openfusion} & 2024 & D: SDF-based & Modular & Object Grounding & \checkmark & Color, Distance field, Seem features \\ \hline
    \rule{0mm}{4mm}SEA-Shine \cite{bezuglyj2024seashine} & 2024 & D: SDF-based & End-2-End & 3D Scene Reconstruction & \checkmark & Pointcloud, Mesh, Semantic labels \\ \hline
    \rule{0mm}{4mm}SVRaster \cite{sun2024svraster} & 2024 & D: Voxel-based & Modular & 3D Scene Reconstruction & \checkmark & RADIO-features, Semantic labels, Color \\ \hline
    \rule{0mm}{4mm}FAST-LIVO2 \cite{zheng2024fast} & 2024 & D: Voxel-based & Without learning & 3D Scene Reconstruction, Vehicle Localization & \checkmark & LiDAR features, Camera features \\ \hline
    \rule{0mm}{4mm}AVLMaps \cite{huang2023avlmaps} & 2023 & D: Voxel-based & Modular & Object Grounding & \checkmark & Visual Localization Features, CLIP Features, LSeg Features, Audio-language Features \\ \hline
    \rule{0mm}{4mm}VLMaps \cite{huang2023vlmaps} & 2023 & D: Voxel-based & Modular & Object Grounding, High-Level Action Planning & \checkmark & Pointcloud projection, Lseg Image-text features,  \\ \hline
    \rule{0mm}{4mm}BEV Scene Graph (BSG) \cite{liu2023bsg} & 2023 & D: Voxel-based & End-2-End & Object Grounding, Indoor Vehicle Navigation & \checkmark & Image features, BEV features, Textual-BEV Grid Scene Graph \\ \hline
        
\end{longtable}
\endgroup

Recent developments include Gaussian splatting methods like M3 \cite{zou2025m3}, SplatTalk \cite{thai2025splattalk}, GaussianGraph \cite{wang2025gaussiangraph}, and OpenGaussian \cite{wu2024opengaussian}, enabling real-time 3D reconstruction and rendering of images and visual-language feature maps from new viewpoints.  

The method GPT4Scene \cite{qi2025gpt4scene} represents maps as Bird's-Eye-View images with labeled objects, significantly improving performance in object detection, question-answering, and high-level planning.  

Non-neural, non-trainable approaches like RTabMap \cite{labbe2019rtabmap} and FAST-LIVO2 \cite{zheng2024fast} focus on multimodal robot localization.  

Neural Implicit Representation methods are used for LLM-based reasoning (SpatialLM \cite{spatiallm2025}, SpatialVLM \cite{chen2024spatialvlm}, LL3DA \cite{chen2024ll3da}) and vehicle navigation (Dynamic Implicit Representations \cite{marza2023dynamicimplicitrep}, Robo-VLN \cite{irshad2021robovln}).  

Table \ref{tab:sparse_static} lists contemporary sparse methods for constructing multimodal 3D maps of static scenes. Few end-to-end models exist here, as they typically train separate neural models for object detection, scene graph generation, and reasoning/answering based on object representations.  

\begingroup\tiny\begin{longtable}[!ht]{p{2.3cm}|p{0.5cm}|p{2.2cm}|p{1.5cm}|p{2.0cm}|p{0.5cm}|p{3.5cm}}
    \caption{Modern Sparse Methods for Constructing Multimodal 3D Maps of Static Scenes\label{tab:sparse_static}}\\
    \hline
        \rule{0mm}{4mm}Method & Year & Scene Representation & Learn & Application & Code & Map Modalities \\ \hline
        \endfirsthead		
        \caption[]{(Continuation)} \\
        \hline
        \rule{0mm}{4mm}Method & Year & Scene Representation & Learn & Application & Code & Map Modalities \\ \hline
        \endhead
        \rule{0mm}{4mm}3DGraphLLM \cite{zemskova20243dgraphllm} & 2024 & S: Knowledge graphs & End-2-End & Object Grounding, Question Answering & \checkmark & Object-based Point cloud features, Image features, semantic relation features \\ \hline
        \rule{0mm}{4mm}ConceptGraphs \cite{gu2024conceptgraphs} & 2024 & S: Knowledge graphs & Modular & Object Grounding & \checkmark & Point cloud, Instance Labels, Captions, CLIP features, 3D Scene Graph \\ \hline
        \rule{0mm}{4mm}BBQ-Deductive \cite{linok2024bbq} & 2024 & S: Knowledge graphs & Modular & Object Grounding & \checkmark & Point cloud, Instance Labels, Captions, EVA2, DINOv2 features, 3D Scene Graph \\ \hline
        \rule{0mm}{4mm}BBQ-CLIP \cite{linok2024bbq} & 2024 & S: Knowledge graphs & Modular & Object Grounding & \checkmark & Point cloud, Instance Labels, EVA2, DINOv2 features \\ \hline
        \rule{0mm}{4mm}SpatialRGPT \cite{cheng2024spatialrgpt} & 2024 & S: Knowledge graphs & Modular & Question Answering  & \checkmark & Image , Pointcloud, Region Masks \\ \hline
        \rule{0mm}{4mm}Open3DSG \cite{koch2024open3dsg} & 2024 & S: Knowledge graphs & Modular & Object Grounding & \checkmark & Image features, Text features, Scene Graph Features \\ \hline
        \rule{0mm}{4mm}3D-HetSGP \cite{ma2024HetSGP} & 2024 & S: Knowledge graphs & Modular & Object Grounding & ~ & PointCloud, Heterogeneous scene graph (different edge types) \\ \hline
        \rule{0mm}{4mm}3D-VLAP \cite{wang20243dvlap} & 2024 & S: Knowledge graphs & Modular & Object Grounding & ~ & Point cloud, CLIP features, 3D scene graph \\ \hline
        \rule{0mm}{4mm}CCL-3DSGG \cite{chen2024ccl} & 2024 & S: Knowledge graphs & Modular & Object Grounding, Question Answering, State Prediction & ~ & CLIP Features, Pointcloud, 3D scene graph features \\ \hline
        \rule{0mm}{4mm}Lang3DSG \cite{koch2024lang3dsg} & 2024 & S: Knowledge graphs & End-2-End & Object Grounding & \checkmark & Pointcloud features, CLIP features, 3D Scene Graph \\ \hline
        \rule{0mm}{4mm}SG-LLM \cite{zhang2024SGLLM} & 2024 & S: Knowledge graphs & Modular & Object Grounding, Indoor Vehicle Navigation & ~ & Object point cloud, Semantic Scene Graph \\ \hline
        \rule{0mm}{4mm}Sg-CityU \cite{sun2024sgcity} & 2024 & S: Knowledge graphs & Modular & Question Answering  & ~ & Point cloud, height of the point, colors, and normals, Objects, Spatial relations \\ \hline
        \rule{0mm}{4mm}SGFormer \cite{lv2024sgformer} & 2024 & S: Knowledge graphs & Modular & Object Grounding & \checkmark & Pointcloud,  Text features,  Object caterories, Spatial Relations \\ \hline
        \rule{0mm}{4mm}SGRec3D \cite{Koch2024SGRec3D} & 2024 & S: Knowledge graphs & End-2-End & Object Grounding, 3D Scene Reconstruction & ~ & Point cloud features, Object categories, Relations (3D scene graph) \\ \hline
        \rule{0mm}{4mm}OVSG \cite{chang2023contextaware3dgraphs} & 2023 & S: Knowledge graphs & Modular & Object Grounding & \checkmark & Image features, Text features, PointCloud Features, Scene Graph Features \\ \hline
        \rule{0mm}{4mm}ViL3DRel \cite{chen2022ViL3DRel} & 2023 & S: Knowledge graphs & End-2-End & Object Grounding & \checkmark & Pointcloud, Text features \\ \hline
        \rule{0mm}{4mm}MonoSSG \cite{wu2023incremental} & 2023 & S: Knowledge graphs & Modular & Object Grounding & \checkmark & Images, Scene graph: node and edge features \\ \hline
        \rule{0mm}{4mm}SMKA \cite{feng2023smka} & 2023 & S: Knowledge graphs & Modular & Object Grounding & \checkmark & Pointcloud, Hierarchical knowledge graph \\ \hline
        \rule{0mm}{4mm}SceneGraphFusion \cite{wu2021scenegraphfusion} & 2021 & S: Knowledge graphs & Modular & Object Grounding & \checkmark & Images, Depths, Dynamic Scene graph: node and edge features \\ \hline
        \rule{0mm}{4mm}SGGpoint \cite{zhang2021sggpoint} & 2021 & S: Knowledge graphs & Modular & Object Grounding & \checkmark & Point cloud features, 3D scene graph \\ \hline
        \rule{0mm}{4mm}3D Scene Graph \cite{armeni20193dscenegraph} & 2019 & S: Knowledge graphs & Modular & Object Grounding & \checkmark & Poit cloud, Object labels, Attributes (Color, Material), Hierarchical Knowledge Graph \\ \hline
        \rule{0mm}{4mm}Chat-Scene \cite{huang2024chatscene} & 2024 & S: Object-based & Modular & Object Grounding, Question Answering & \checkmark & Object point cloud, image masks, image-text (DINOv2) features, point cloud-text (Uni3D) features \\ \hline
        \rule{0mm}{4mm}Grounded 3D LLM \cite{chen2024grounded3dllm} & 2024 & S: Object-based & Modular & Object Grounding, Question Answering & \checkmark & Pointcloud, color, pointcloud-text features, 3D mask features \\ \hline
        \rule{0mm}{4mm}HOV-SG \cite{werby2024hierarchicalov3d} & 2024 & S: Object-based with Hierarchy & Modular & Object Grounding & \checkmark & Color Pointcloud, Hierarhical 3D Scene graph, CLIP features \\ \hline
        \rule{0mm}{4mm}CURB-SG \cite{greve2024curbcs} & 2024 & S: Object-based with Hierarchy & Modular & Vehicle Localization & \checkmark & Roads \& intersections, Landmarks, Lane graph \& vehicles, Keyframes \& point clouds \\ \hline
        \rule{0mm}{4mm}Opengraph \cite{deng2024opengraph} & 2024 & S: Object-based with Hierarchy & Modular & Object Grounding, Path Planning,  & \checkmark & Segments, Instances, Lane Graph, Point Cloud \\ \hline
        \rule{0mm}{4mm}MSG \cite{zhang2024MSG} & 2024 & S: Object-based with Hierarchy & End-2-End & Object Grounding, Place Recognition & \checkmark & Place embedding, Object Embedding \\ \hline
        \rule{0mm}{4mm}Point2Graph \cite{xu2024point2graph} & 2024 & S: Object-based with Hierarchy & Modular & Object Grounding, Indoor Vehicle Navigation & \checkmark & Pointcloud, Color, CLIP Features, Instance Labels, Hieararchical 3D Scene Graph \\ \hline
        \rule{0mm}{4mm}Hydra \cite{hughes2024foundationsofspatialperception} & 2023 & S: Object-based with Hierarchy & Modular & Object Grounding, Place Recognition & \checkmark & 3D Mesh, Place Descriptor, Hierarchical 3D Scene Graph, Semantic labels \\ \hline
        \rule{0mm}{4mm}SayPlan \cite{rana2023sayplan} & 2023 & S: Object-based with Hierarchy & Modular & Action Planning & \checkmark & Hierarchical Graph \\ \hline
        \rule{0mm}{4mm}DSG-RL \cite{ravichandran2022dsgrl} & 2022 & S: Object-based with Hierarchy & Modular & Vehicle Navigation & \checkmark & Hierarchical Navigation Graph, Mesh, Object labels, Place features \\ \hline
        \rule{0mm}{4mm}Spatial Ontologies \cite{strader2024indoorontologies} & 2024 & S: Ontology-based & Modular & Object Grounding & ~ & Hierarchical 3D Scene Graph (Mesh, Objects, Places, Regions) \\ \hline
\end{longtable}
\endgroup

Few methods address vehicle localization (CURB-SG \cite{greve2024curbcs}, MSG \cite{zhang2024MSG}, Hydra \cite{hughes2024foundationsofspatialperception}) or robot navigation (SG-LLM \cite{zhang2024SGLLM}, DSG-RL \cite{ravichandran2022dsgrl}) and path planning (Opengraph \cite{deng2024opengraph}, SayPlan \cite{rana2023sayplan}).  

More methods perform question-answering using LLMs, such as 3DGraphLLM \cite{zemskova20243dgraphllm}, Chat-Scene \cite{huang2024chatscene}, SpatialRGPT \cite{cheng2024spatialrgpt}, CCL-3DSGG \cite{chen2024ccl}, and Sg-CityU \cite{sun2024sgcity}, with the latter addressing city-scale scene questions.  

Most methods in Table \ref{tab:sparse_static} generate scene graphs and thus address Object Grounding. Overall, existing methods for static 3D scenes lack universality due to the absence of unified benchmarks covering major applications.  

\section{M3DMap Method}  
\label{sec:proposed_approach}  

\begin{figure}[!ht]  
  \centering  
  \includegraphics[width=0.99\textwidth]{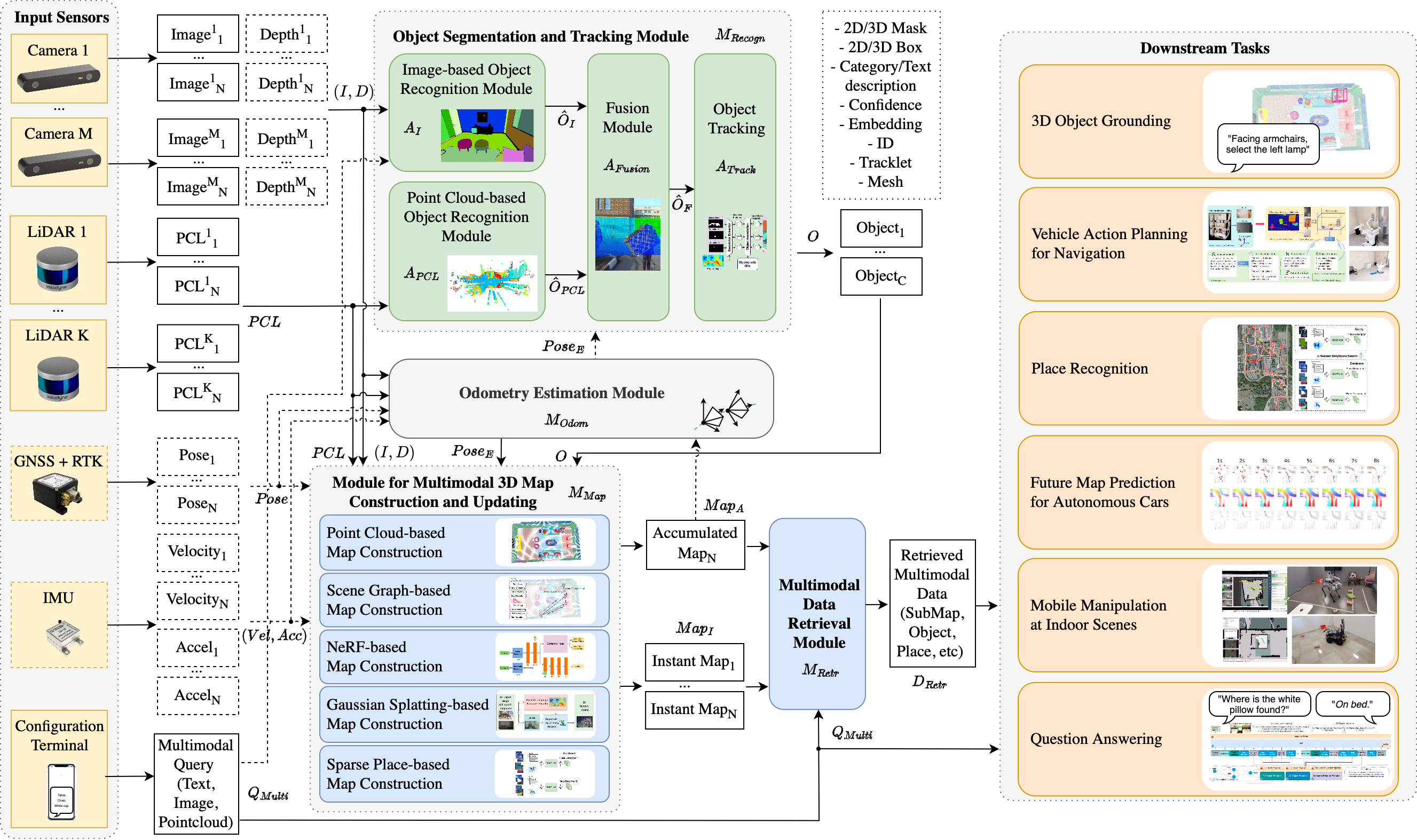}  
  \caption{Schematic of the proposed modular M3DMap approach for Object-aware Multimodal 3D Mapping, including Dynamic Environments}  
  \label{fig:m3dmap}  
\end{figure}

This work proposes a modular method for constructing multimodal 3D maps, M3DMap (Multimodal 3D Mapping), suitable for dynamic environments. Its general scheme is shown in Figure \ref{fig:m3dmap}.  

The method assumes the following input sensors:  
\begin{itemize}  
    \item One or more ($M$) RGB-D or stereo cameras, producing sequences of $N$ images $I$ and depth maps $D$.  
    \item One or more ($K$) LiDARs, generating point cloud sequences $PCL$.  
    \item A Global Navigation Satellite System with RTK corrections (GNSS+RTK) for outdoor localization, producing robot pose sequence $Pose$. This is optional, as satellite data may be unavailable or inaccurate indoors or in urban canyons.  
    \item An Inertial Measurement Unit (IMU) estimating robot velocity $Vel$ and acceleration $Acc$. This is also optional due to noise susceptibility.  
    \item A configuration terminal for inputting multimodal queries (text, images, or point clouds) $Q_{Multi}$ to modules constructing or using maps for downstream tasks.  
\end{itemize}  

Object-aware multimodal 3D mapping, yielding an accumulated map $Map_A$ at step $N$ and instantaneous maps $Map_I$ at each time step, is formally expressed as:  

\begin{equation}  
    (Map_A, Map_I)=\mathbb{M}(I,D,PCL,[Pose],[Vel],[Acc],[Q_{Multi}]),  
\end{equation}  
where $\mathbb{M}$ is the modular method comprising three key interconnected modules, described below. Brackets $[.]$ denote optional inputs, which may or may not be used in different implementations.  

The first module, Object Segmentation and Tracking, processes sensor data to generate a set $O$ of $C$ objects:  

\begin{equation}  
    O=M_{Recogn}(I,D,PCL,[Pose_E],[Q_{Multi}]),  
\end{equation}  
where $Pose_E$ is the optional robot pose estimated by the Odometry Estimation Module. Explicit pose estimation is typically needed for point cloud densification, object tracking, and training on data sequences.  

The Odometry Estimation Module is formally described as:  
\begin{equation}  
    Pose_E=M_{Odom}(I,D,PCL,[Pose],[Vel],[Acc],[Map_A]).  
\end{equation}  

The third module, Multimodal 3D Map Construction and Updating, has implementations varying by the desired map representation but is formally unified as:  
\begin{equation}  
    (Map_A, Map_I)=M_{Map}(I,D,PCL,O,Pose_E,[Pose],[Vel],[Acc]).  
\end{equation}  
Generating instantaneous maps $Map_I$ accounts for scene dynamics, storing up-to-date maps $Map_{Ii}$ at key timesteps for querying.  

For downstream tasks, retrieving multimodal data $D_{Retr}$ (submaps, target objects, places, etc.) from maps is crucial. This is handled by the Multimodal Data Retrieval Module:  
\begin{equation}  
    D_{Retr}=M_{Retr}(Map_A, Map_I, Q_{Multi}).  
\end{equation}  

Details of all modules are provided in Subsections \ref{sec:seg_module}–\ref{sec:retrieve_module}.  

\subsection{Object Segmentation and Tracking Module}  
\label{sec:seg_module}  
This module includes Image-based Object Recognition Module and Point Cloud-based Object Recognition Module, which operate independently and generate preliminary results of object recognition based on data of different modalities.

The first of them implements an algorithm for analyzing a sequence of stereo camera data with preliminary visual recognition of objects $\hat{O}_{I}$
\begin{equation}
\hat{O}_{I}=A_{I}(I, D, [Q_{Multi}]).
\end{equation}
In this case, recognition can be performed either with a fixed set of categories (as in FCNResNet-MOC \cite{shepel2021occupancygrid}, DAGM-Mono \cite{murhij2024dagmmono}, Center3dAugNet \cite{belkin2022center3daugnet}), or with an open dictionary (as in the SiB \cite{avshalumov2025sib}, Reframing \cite{avshalumov2024reframing}, TASFormer \cite{yudin2023tasformer} approaches) or using visual queries $Q_{Multi}$ (as in the STRL-Robotics framework \cite{mironov2024strl}). The result of recognition is either 2D or 3D bounding rectangles or masks, depending on the algorithm implementation. In addition, the vector $\hat{O}_{I}$ can also include vector representations of objects using CLIP, DINOv2, and a text description of the objects, as implemented in \cite{linok2024bbq}.

The second module implements an algorithm for preliminary recognition of objects $\hat{O}_{PCL}$ from a sequence of point clouds from LiDARs
\begin{equation}
\hat{O}_{PCL}=A_{PCL}(PCL, [Pose_E]).
\end{equation}
It can use the pose estimate $Pose_E$ to compact the input point cloud (as in FMFNet \cite{murhij2022fmfnet} and RVCDet \cite{murhij2022rvcdet}), which, as a rule, improves the quality of detection and segmentation of objects, especially those located far from the robot or vehicle.
The result of the algorithm recognition is 3D bounding rectangles or masks (as in DAPS3D \cite{klokov2023daps3d}) depending on the implementation. In addition, the $\hat{O}_{PCL}$ vector can also include 3D vector representations of objects, for example, using Uni3D (as in 3DGraphLLM \cite{zemskova20243dgraphllm}), MinkLoc3Dv2 (as in MSSPlace \cite{melekhin2024mssplace}).

The Fusion module allows to combine preliminary recognition results (segmentation or detection) $\hat{O}_{F}$ from both modules, taking into account their mutual calibration data (as in DAPS3D \cite{klokov2023daps3d})
\begin{equation}
\hat{O}_{F}=A_{Fusion}(\hat{O}_{I}, \hat{O}_{PCL}).
\end{equation}

This sequence of preliminary recognition results is fed to the Object Tracking Module, which implements various association algorithms based on the Hungarian algorithm and ensures that objects are assigned a unique identifier and tracklets with their previous movement trajectory (as in \cite{basharov2021tracking}).
\begin{equation}
O=A_{Track}(\hat{O}_{F}, [Pose_E]).
\end{equation}
In this module, when performing tracking in three-dimensional space, the cost function includes, among other things, the intersection of objects over the region, taking into account the 6DoF movement of the robot, determined on the basis of the estimated pose $Pose_E$ (as in the work of FMFNet \cite{murhij2022fmfnet}).

\subsection{Odometry Estimation Module}  
\label{sec:odom_module}  
This module is based on SLAM (Simultaneous Localization and Mapping) optimization approaches. Two main implementations of this module have been developed to work in dynamically changing environments:
\begin{itemize}
\item An approach to estimating poses from pre-built lidar maps with the elimination of point clouds belonging to dynamic objects (as in the work of \cite{belkin2024lmrinprior}). Point clouds are filtered based on masks determined by image segmentation using deep learning models, such as ResnetFCN-MOC \cite{shepel2021occupancygrid} or the multi-task model \cite{basharov2022multitask} .
\item An approach to integrating navigation data from a GNSS+RTK system and lidar odometry data for operation in high-rise buildings/underground areas where satellite navigation is inaccurate and in desert areas where lidar localization errors are possible (as in the works of AnKF \cite{ladanova2023AnKF} and \cite{abdrazakov2021ukfneural}).
\end{itemize}

\subsection{Module for 3D Map Construction and Updating}  
\label{sec:map_module}  
This module implements algorithms for constructing various representations of multimodal 3D maps depending on the problem being solved. Among them:
\begin{itemize}
\item Point Cloud/Voxel-based Map Construction (as in LMR in Prior Maps \cite{belkin2024lmrinprior}, DAPS3D \cite{klokov2023daps3d}, Semantic Occupancy Mapping \cite{shepel2021occupancygrid}).
\item Scene Graph-based Map Construction (as in BBQ approaches \cite{linok2024bbq}, 3DGraphLLM \cite{zemskova20243dgraphllm}, LAMDEN \cite{zhang2025elevnav}).
\item NeRF-based Map Construction (as in uSF \cite{skorokhodov2024usf}, NeRFUS \cite{zubkov2025nerfus}).
\item Gaussian Splatting-based Map Construction (as in LEG-SLAM \cite{titkov2025legslam}).
\item Sparse Place-based Map Construction (as in HPointLoc \cite{yudin2022hpointloc}, MSSPlace \cite{melekhin2024mssplace}).
\end{itemize}

\subsection{Multimodal Data Retrieval Module}  
\label{sec:retrieve_module}  
Two implementation options of this module have been studied and have proven their effectiveness:
\begin{itemize}
\item Implementation based on multimodal data encoders (image, point cloud, semantic masks, text), which allows fast searching in a sparse map in the form of a database of individual places (as in the MSSPlace method \cite{melekhin2024mssplace}).
\item Algorithm based on a large language model (LLM), which allows selecting a preliminary list of target objects related to the requested object from a text representation of the scene graph (as in the BBQ method \cite{linok2024bbq}).
\end{itemize}

\section{Some Theoretical Background for Combining Various Data Sources in the Proposed Architecture}  
\label{sec:theory}  

This section provides theoretical justification for the benefits of using multiple modalities in the Object Segmentation and Tracking Module and applying additional data processing alongside raw sensor inputs in the Module for 3D Map Construction and Updating (see Figure \ref{fig:m3dmap}).  

\begin{figure}[!ht]  
  \centering  
  \includegraphics[width=0.85\textwidth]{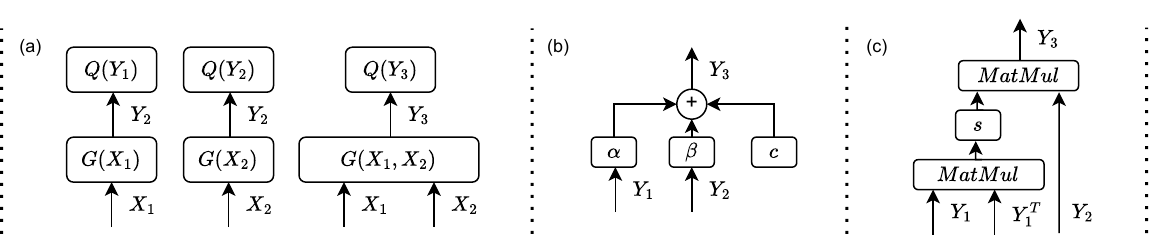}  
  \caption{Explanation of the proposed propositions on combining modalities $y_1$ and $y_2$ to improve model output $y_3$: (a) general setup, (b) linear fusion, (c) matrix attention-like fusion}  
  \label{fig:theory}  
\end{figure}  

Let $Y_1 = G_1(X_1)$ and $Y_2 = G_2(X_2)$ be the results of two recognition algorithms for input data tensors $X_1$ and $X_2$, synchronized in time, which may represent data of different modalities, for example, point clouds $P$ or images $I$, or the output of a nonlinear model processing such data. These two sources are fed to the model $G(X_1,X_2)$, the output of which is the vector $Y_3$, where each element of the output vectors $Y$ is in the range $y\in[0,1]$ and is the degree of classification confidence, which corresponds, for example, to the problem of recognizing objects on a map based on input queries. The quality of this output signal is then estimated by the quality metric $Q(Y_3)$. This setup is shown schematically in Figure \ref{fig:theory}a.

Let us prove under what conditions the quality of recognition of $Q(Y_3)$, which processes two modalities at once, is higher than the quality of recognition of the algorithms processing each modality separately $Q(Y_1)$ and $Q(Y_2)$. That is, the system of inequalities must be satisfied:
\setcounter{equation}{9}
\begin{equation}\label{eq:quality}
\begin{cases}
Q(Y_3) > Q(Y_1),
\\
Q(Y_3) > Q(Y_2).
\end{cases}
\space
\Rightarrow
2Q(Y_3) > Q(Y_1)+Q(Y_2).
\end{equation}

For the right and left parts of the equation, we can take the mathematical expectation, which will not change the inequality:

\begin{equation}\label{eq:e_quality}
2\mathbb{E}[Q(Y_3)] > \mathbb{E}[Q(Y_1)]+\mathbb{E}[Q(Y_2)].
\end{equation}

Let the quality metric $Q$ be defined as the completeness of the classification by the formula

\begin{equation}\label{eq:e_metric}
Q(Y)=\frac{\sum^{N_{Y}}_1[y_{i}\geq t]}{N_{Y}},
\end{equation}
where $t>0$ is some threshold, $N_{Y}$ is the amount of data in the sequence $Y$, $[.]$ is the designation of the indicator function.

Then
\begin{equation}\label{eq:prob_quality} 
\mathbb{E}[Q(Y)] = \frac{\mathbb{E}[\sum^{N_{Y}}_1[y_{i}\geq t]]}{N_{Y}} 
= \frac{\sum^{N_{Y}}_1\mathbb{E}[[y_{i}\geq t]]}{N_{Y}} 
= \frac{N_{Y}\mathbb{E}[Y\geq t]}{N_{Y}}=\mathbb{E}[Y\geq t]=P(Y\geq t),
\end{equation}
where $P$ is the probability.

Based on Markov's inequality, it is known that
\begin{equation}\label{eq:prob_quality}
P(Y\geq t)\leq \frac{\mathbb{E}[Y]}{t},
\end{equation}

Using this inequality, we can rewrite inequality (\ref{eq:e_quality}) as
\begin{equation}
2\frac{\mathbb{E}[Y_3]}{t} \geq 2\mathbb{E}[Q(Y_3)] > \frac{\mathbb{E}[Y_1]}{t}+\frac{\mathbb{E}[Y_2]}{t} \geq \mathbb{E}[Q(Y_1)]+\mathbb{E}[Q(Y_2)].
\end{equation}

From here we obtain
\begin{equation}\label{eq:e_fusion}
2\mathbb{E}[Y_3] > \mathbb{E}[Y_1]+\mathbb{E}[Y_2].
\end{equation}

Now let us consider two special cases of how the models generating the outputs $Y_i$ shown in Figure~\ref{fig:theory} can be related to each other.

\textbf{First case.} In this case, there is a linear relationship between the models (see Figure \ref{fig:theory}b):
\begin{equation}\label{eq:linear_fusion}
Y_3 = \alpha Y_1 + \beta Y_2 + c.
\end{equation}

Substituting the expression (\ref{eq:linear_fusion}) into (\ref{eq:e_fusion}) yields:
\begin{equation}
2\mathbb{E}[\alpha Y_1 + \beta Y_2 + c] > \mathbb{E}[Y_1]+\mathbb{E}[Y_2],
\end{equation}

\begin{equation}
2\alpha \mathbb{E}[Y_1] + 2\beta \mathbb{E}[Y_2] + 2c > \mathbb{E}[Y_1]+\mathbb{E}[Y_2],
\end{equation}

\begin{equation}\label{eq:linear_theorem}
(2\alpha - 1) \mathbb{E}[Y_1] + (2\beta-1) \mathbb{E}[Y_2] + 2c > 0,
\end{equation}

Since from the condition $\mathbb{E}[Y_1]\geq 0$ and $\mathbb{E}[Y_2]\geq 0$, then the inequality (\ref{eq:linear_theorem}) is satisfied with the true value of the following condition, which can be satisfied as a result of training the classification model $G$:
\begin{equation}\label{eq:linear_condition}
\mathbb{C}_{case_1} = (\alpha > 0.5) \And (\beta > 0.5) \And (c>0).
\end{equation}

\textbf{The second case.} In this case, there is an attention-like connection between the models with two matrix products (see Figure \ref{fig:theory}c):
\begin{equation}\label{eq:attention_fusion}
Y_3 = (s Y_1 \times Y^T_1 )\times Y_2.
\end{equation}

We substitute the expression (\ref{eq:attention_fusion}) into (\ref{eq:e_fusion}) and obtain:
\begin{equation}\label{eq:attention_eq}
2\mathbb{E}[(s Y_1 \times Y^T_1 )\times Y_2] > \mathbb{E}[Y_1]+\mathbb{E}[Y_2],
\end{equation}

It is known that the mathematical expectation of the product of two matrix functions is
\begin{equation}\label{eq:matrix_product}
\mathbb{E}[A(X)B(X)] = \mathbb{E}[A(X)] \mathbb{E}[B(X)] + Cov(A(X), B(X)),
\end{equation}
where $ Cov(A(X), B(X))$ is a covariance matrix, each element of which is the sum of the covariances of the elements A and B involved in calculating the corresponding element of the product.

Taking into account the equation (\ref{eq:matrix_product}) we obtain on the basis of (\ref{eq:attention_eq}):
\begin{equation}
2\mathbb{E}[s Y_1 \times Y^T_1] \mathbb{E}[Y_2] + 2Cov(s Y_1 \times Y^T_1, Y_2) > \mathbb{E}[Y_1]+\mathbb{E}[Y_2],
\end{equation}

\begin{equation}
2s(\mathbb{E}[Y_1] \mathbb{E}[Y_1] + Cov(Y_1, Y_1)) \mathbb{E}[Y_2] + 2Cov(s Y_1 \times Y^T_1, Y_2) > \mathbb{E}[Y_1]+\mathbb{E}[Y_2],
\end{equation}

\begin{equation}
2s\mathbb{E}[Y_1] \mathbb{E}[Y_1]\mathbb{E}[Y_2] + 2sCov(Y_1, Y_1) \mathbb{E}[Y_2] + 2Cov(s Y_1 \times Y^T_1, Y_2) > \mathbb{E}[Y_1]+\mathbb{E}[Y_2],
\end{equation}

\begin{equation}\label{eq:attention_theorem}
\mathbb{E}[Y_1](2s\mathbb{E}[Y_1]\mathbb{E}[Y_2]-1) + \mathbb{E}[Y_2] (2sCov(Y_1, Y_1)-1) + 2Cov(s Y_1 \times Y^T_1, Y_2) > 0,
\end{equation}

Note that the covariance can be calculated as $Cov(A, B)=\rho_{A,B}\sigma_{A}\sigma_{B}$, where $\sigma_{A}$ and $\sigma_{B}$ are the standard deviations, $\rho_{A,B}$ is the Pearson correlation coefficient, which is positive when the two functions are directly dependent and negative when they are inversely dependent (that is, when, for example, the function $B$ increases as $A$ decreases), it is also equal to one if $A = B$ (in which case the covariance becomes the variance).

Since from the condition $\mathbb{E}[Y_1]\geq 0$ and $\mathbb{E}[Y_2]\geq 0$, then the inequality (\ref{eq:attention_theorem}) is satisfied under two conditions simultaneously, which can be achieved as a result of training the classification model $G$:

1) $(s > 1/(\mathbb{E}[Y_1]\mathbb{E}[Y_2])+1/(\sigma^2_{Y_1}))$, the constraint on the scale factor $s$, which is obtained from the first two terms of the inequality (\ref{eq:attention_theorem}):

- $(s > 1/(2\mathbb{E}[Y_1]\mathbb{E}[Y_2]))$, that is, there must be a constraint on the scale factor $s$ based on the mathematical expectations of the models $Y_1$ and $Y_2$,

- $(s > 1/(2\sigma^2_{Y_1}))$, that is, there is also a limitation on the scale factor $s$ taking into account the dispersion of the model $Y_1$,

2)$(\rho_{s Y_1 \times Y^T_1, Y_2}>0)$, that is, with the growth of the matrix product $Y_1 \times Y^T_1$, $Y_2$ should also grow.

This can be written as a single condition that must be true for the inequality (\ref{eq:attention_theorem}) to hold:
\begin{equation}\label{eq:attention_condition}
\mathbb{C}_{case_2} = (s > 1/(\mathbb{E}[Y_1]\mathbb{E}[Y_2])+1/(\sigma^2_{Y_1})) \And (\rho_{s Y_1 \times Y^T_1, Y_2}>0).
\end{equation}

Thus, the above arguments prove the following two theorems.

\textbf{Theorem 1.} \textit{There are recognition models $Y_1=G_1(X_1)$, $Y_2=G_2(X_2)$, $Y_3=G(X_1,X_2)$, for which the recognition quality of model $Q(Y_3)$, which processes two input signals $X_1$,$X_2$ at once, is higher than the recognition quality of algorithms processing each signal separately $Q(Y_1)$ and $Q(Y_2)$. The recognition quality of models $Q(Y)$ is determined by the formula (\ref{eq:e_metric}).}

This theorem yields two important consequences.

\textbf{Corollary 1.} \textit{There are such multimodal recognition models $Y_3=G(X_1,X_2)$ for which the recognition quality of the model $Q(Y_3)$, which processes two input modalities $X_1$,$X_2$ at once, for example, images and point clouds, is higher than the recognition quality of the algorithms for processing $Y_1=G_1(X_1)$,$Y_2=G_2(X_2)$, each modality separately $Q(Y_1)$ and $Q(Y_2)$.}

This corollary shows that the simultaneous use of several different modalities (image, point cloud, text) allows to improve the quality of object recognition, which is an integral part of the M3DMap multimodal map construction system.

\textbf{Corollary 2.} \textit{There are such recognition models $Y_3=G(X,F(X))$ for which the recognition quality of the model $Q(Y_3)$, which processes two input sources of the signal $X$,$F(X)$ at once, for example, images and their features, is higher than the recognition quality of the algorithms for processing $Y_1=G_1(X)$,$Y_2=G_2(F(X))$, each of the signals separately $Q(Y_1)$ and $Q(Y_2)$.}

This corollary reflects the benefit of the simultaneous use of raw data and additional feature encoders, including those based on fundamental models, to improve the quality of object recognition when forming and using various representations of multimodal 3D maps.

\textbf{Theorem 2.} \textit{If the relationship between the recognition models $Y_1=G_1(X_1)$, $Y_2=G_2(X_2)$, $Y_3=G(X_1,X_2)$ is specified in the form (\ref{eq:linear_fusion} and (\ref{eq:attention_fusion}) and the conditions (\ref{eq:linear_condition}) and (\ref{eq:attention_condition}) are true, then the recognition quality of the model $Q(Y_3)$, determined by the formula (\ref{eq:e_metric}), is higher than the recognition quality of $Q(Y_1)$ and $Q(Y_2)$ for each of the models separately.}

The second theorem theoretically shows the high applicability of linear layers based on multilayer perceptrons (MLP) and attention and cross-attention modules for constructing modern high-quality multimodal architectures models that are designed to solve various practical problems related to the recognition (classification, detection, segmentation, search, etc.) of objects on three-dimensional maps.







\section{The Effect of Using Different Architecture Options in Downstream Task Solving}  
\label{sec:effect}  

This section presents the experimental results demonstrating the advantages of the approaches used in constructing the main modules of the proposed M3DMap method.

\textbf{ Object segmentation and tracking module.} 
Table \ref{tab:exp_seg_track} shows the results of detection and segmentation of scene objects in images on GOLD-A \cite{avshalumov2024reframing} and RefCOCOg \cite{kazemzadeh2014refcocog} Datasets, respectively, using the developed methods for training neural network models Reframing \cite{avshalumov2024reframing} (based on the PPO reinforcement learning method) and SiB \cite{avshalumov2025sib} (based on the GRPO method) to improve text queries fed to the input of popular models GroundingDINO \cite{liu2024groundingdino}, YOLO-World \cite{cheng2024yoloworld}, YOLO-E \cite{wang2025yoloe}. The table shows that for all models there is a significant increase in recognition quality, which indicates the potential for using such approaches in cases where detection or segmentation of objects by a text query is required.

Also, Table \ref{tab:exp_seg_track} contains the results of Zero-shot 3D semantic segmentation by a text query on the ScanNet \cite{dai2017scannet} dataset. It follows from it that the developed BBQ-CLIP \cite{linok2024bbq} method, which uses an advanced approach to encoding scene objects using the EVA variety of CLIP and object tracking based on DINOv2 features, allows for significantly more accurate implementation of 3D scene segmentation than its closest analogues. 

\begin{table}[!ht]
    \caption{Quality of multimodal learnable object segmentation and tracking}
    \label{tab:exp_seg_track}
    \centering
    \tiny
    \begin{tabular}{l|l|l}
    \hline
        \rule{0mm}{4mm}Recognition model & Text prompt tuning approach & Metrics, mIoU \\ \hline
        \multicolumn{3}{l}{\rule{0mm}{4mm}GOLD-A Dataset \cite{avshalumov2024reframing} (Object detection)}\\ \hline
        \rule{0mm}{4mm}GroundingDINO \cite{liu2024groundingdino} & Original prompt & 0.572 \\ \hline
        \rule{0mm}{4mm}GroundingDINO \cite{liu2024groundingdino} & Reframing(Llama2-7b) \cite{avshalumov2024reframing} & 0.585 \\ \hline
        \rule{0mm}{4mm}GroundingDINO \cite{liu2024groundingdino} & SiB (Llama2-7b) \cite{avshalumov2025sib} & \textbf{0.593} \\ \hline
        \rule{0mm}{4mm}YOLO-World \cite{cheng2024yoloworld} & Original prompt & 0.535 \\ \hline
        \rule{0mm}{4mm}YOLO-World \cite{cheng2024yoloworld} & Reframing(Llama2-7b) \cite{avshalumov2024reframing}  & 0.545 \\ \hline
        \rule{0mm}{4mm}YOLO-World \cite{cheng2024yoloworld} & SiB (Llama2-7b) \cite{avshalumov2025sib} & \textbf{0.550 }\\ \hline
        \multicolumn{3}{l}{\rule{0mm}{4mm}RefCOCOg Dataset \cite{kazemzadeh2014refcocog} (Object segmentation)} \\ \hline
        \rule{0mm}{4mm}YOLO-E \cite{wang2025yoloe} & Original prompt  & 0.276 \\ \hline
        \rule{0mm}{4mm}YOLO-E \cite{wang2025yoloe} & Pretrained model (Qwen2.5-7b)  & 0.232 \\ \hline
        \rule{0mm}{4mm}YOLO-E \cite{wang2025yoloe} & SiB (Qwen2.5-7b) \cite{avshalumov2025sib} & \textbf{0.395} \\ \hline
        \multicolumn{3}{l}{\rule{0mm}{4mm}ScanNet \cite{dai2017scannet} (Zero-shot 3D semantic segmentation)}\\ \hline
        \rule{0mm}{4mm}ConceptFusion \cite{jatavallabhula2023conceptfusion} & Original object label & 0.26 \\ \hline
        \rule{0mm}{4mm}OpenMask3D \cite{takmaz2023openmask3d} & Original object label & 0.18 \\ \hline
        \rule{0mm}{4mm}ConceptGraphs \cite{gu2024conceptgraphs} & Original object label & 0.26 \\ \hline
        \rule{0mm}{4mm}BBQ-CLIP \cite{linok2024bbq} & Original object label & \textbf{0.34} \\ \hline
    \end{tabular}
\end{table}

\textbf{Odometry Estimation Module.} 
Table \ref{tab:exp_odometry} shows two important positive effects that are observed when using different implementations of the Odometry Estimation Module. For the implementation option in the form of the UKF (PNC+MNC CNN) algorithm, which combines navigation data from the OpenVSLAM visual odometry algorithms \cite{sumikura2019openvslam} and the A-LeGO-LOAM lidar odometry \cite{legoloam2018} based on a modified trainable Kalman filter, a significant reduction in the localization error by position and rotation angle is observed compared to the quality estimates of the navigation data from each source separately.

For the second version of the module implementation based on the LMR in prior maps method \cite{belkin2024lmrinprior}, a significant reduction in localization errors by position and angle is shown in the case of taking into account the masks of dynamic objects identified by the FCNResNet-MOC neural network image segmentation model \cite{shepel2021occupancygrid}.

\begin{table}[!t]
    \caption{Impact of Using Different Modalities on the Odometry Estimation Module}
    \label{tab:exp_odometry}
    \centering
    \tiny
    \begin{tabular}{l|l|p{1.0cm}|p{1.0cm}|p{1.1cm}}
    \hline
        \rule{0mm}{4mm}Method & Modalities & Semantics usage & Trans. error [m] & Rot. error [deg] \\ \hline
        \multicolumn{5}{l}{\rule{0mm}{4mm}SDBCS Husky fusion dataset \cite{abdrazakov2021ukfneural} (robot odometry estimation and localization)} \\ \hline
        \rule{0mm}{4mm}OpenVSLAM \cite{sumikura2019openvslam} & RGB, Depth, Visual Point Cloud & No & 0.371 & 0.674 \\ \hline
        \rule{0mm}{4mm}A-LeGO-LOAM \cite{legoloam2018} & LiDAR Point Cloud & No & 0.114 & 0.706 \\ \hline
        \rule{0mm}{4mm}UKF \cite{wan2000ukf} & RGB, Depth, Visual and LiDAR Point Cloud & No & 0.114 & 0.521 \\ \hline
        \rule{0mm}{4mm}UKF (PNC+MNC CNN) \cite{abdrazakov2021ukfneural} & RGB, Depth, Visual and LiDAR Point Cloud & No & \textbf{0.105 }& \textbf{0.484 }\\ \hline
        \multicolumn{5}{l}{\rule{0mm}{4mm}SDBCS Husky full dataset \cite{belkin2024lmrinprior} (robot odometry estimation and localization)} \\ \hline
        \rule{0mm}{4mm}ORB-SLAM2 \cite{mur2017orbslam2} & RGB, Depth, Point Cloud & No & 0.292 & 0.636 \\ \hline
        \rule{0mm}{4mm}GMMLoc \cite{huang2020gmmloc} & RGB, Gaussian Mixture Model & No & 0.289 & 0.715 \\ \hline
        \rule{0mm}{4mm}Iris (RGBD) \cite{yabuuchi2021iris} & RGB, Depth, Point Cloud & No & 0.387 & 2.775 \\ \hline
        \rule{0mm}{4mm}LOL \cite{zhang2017lol} & RGB, Depth, LiDAR Point Cloud & No & 0.138 & 0.644 \\ \hline
        \rule{0mm}{4mm}LMR in prior maps (basic) \cite{belkin2024lmrinprior} & RGB, Depth, LiDAR Point Cloud & No & 0.166 & 0.553 \\ \hline
        \rule{0mm}{4mm}LMR in prior maps (+ odometry) \cite{belkin2024lmrinprior} & RGB, Depth, LiDAR Point Cloud & No & 0.132 & 0.539 \\ \hline
        \rule{0mm}{4mm}LMR in prior maps (+ odometry + multiframe) \cite{belkin2024lmrinprior}  & RGB, Depth, LiDAR Point Cloud & No & 0.125 & 0.542 \\ \hline
        \rule{0mm}{4mm}LMR in prior maps (+ odometry + multiframe + mask) \cite{belkin2024lmrinprior} & RGB, Depth, LiDAR Point Cloud & Yes & \textbf{0.115} & \textbf{0.419} \\ \hline
    \end{tabular}
\end{table}

\textbf{Module for 3D Map Construction and Updating.} 
Tables \ref{tab:exp_point_map} - \ref{tab:exp_nerf_splat} show the results of experiments demonstrating the implementation options of the Module for 3D map construction and updating with different ways of representing a 3D scene.

Table \ref{tab:exp_point_map} shows the effect of constructing semantic maps of passability based on voxels on the open Semantic KITTI dataset \cite{behley2019semantickitti}. It is shown that using the results of segmentation of dynamic objects (people, vehicles) can significantly improve obstacle classification precision, which is important for increasing the safety of mobile robots in a changing environment.

\begin{table}[!t]
    \caption{Quality of multimodal voxel-based map construction on Semantic KITTI Dataset \cite{behley2019semantickitti} for 3D occupancy map reconstruction and semantic segmentation tasks}
    \label{tab:exp_point_map}
    \centering
    \tiny
    \begin{tabular}{l|l|l}
    \hline
        \rule{0mm}{4mm}Method & Modalities & Obstacle classification precision, \% \\ \hline
        \rule{0mm}{4mm}Occupancy Mapping (Raw point cloud) \cite{shepel2020modifiedoccupancy} & Image, Visual Point Cloud & 50.2 \\ \hline
        \rule{0mm}{4mm}Semantic Occupancy Mapping (FCNResNetMOC) \cite{shepel2021occupancygrid} & Image, Visual Point Cloud, Semantic labels & \textbf{67.9} \\ \hline
    \end{tabular}
    
\end{table}

Table \ref{tab:exp_scene_graph} shows the significant role played by semantic and spatial relations between objects of the 3D scene graph. It can be seen how the baseline Chat-Scene \cite{huang2024chatscene} model, using only object encoders, performs worse on the Object Grounding task on the ScanRefer dataset than the proposed 3DGraphLLM \cite{zemskova20243dgraphllm} model using the same large Vicuna-7B-v1.5 language model. In addition, the table shows the prospects for using and developing the GPT4Scene-HDM \cite{qi2025gpt4scene} approach, which uses BEV image with object labels (marks), significantly outperforming other approaches on the scene description task on the Scan2Cap dataset.

\begin{table}[!t]
    \caption{Quality of object grounding, question answering, scene captioning using multimodal scene graph-based map}
    \label{tab:exp_scene_graph}
    \centering
    \tiny
    \begin{tabular}{l|p{2.5cm}|l|l|l|l|l|l|l}
    \hline
         & &  & ScanRefer &  & Scan2Cap & & ScanQA &  \\
        Methods & Scene Graph-based representation & LLM & A@0.25↑ & A@0.5↑ & CIDEr@0.5↑ & B-4@0.5↑ & CIDEr↑ & B-4↑ \\ \hline
        \rule{0mm}{4mm}Grounded 3D LLM \cite{chen2024grounded3dllm} & Objects with 3D features & Tiny-Vicuna-1B & 47.9 & 44.1 & 70.6 & 35.5 & 72.7 & 13.4 \\ \hline
        \rule{0mm}{4mm}Chat-Scene \cite{huang2024chatscene} & Objects with 2D, 3D features & Vicuna-7B-v1.5 & 55.5 & 50.2 & 77.1 & 36.3 & 87.7 & 14.3 \\ \hline
        \rule{0mm}{4mm}3DGraphLLM \cite{zemskova20243dgraphllm} & Objects with 2D, 3D features, Semantic and Spatial Edges & Vicuna-7B-v1.5 & 58.6 & 53.0 & 79.2 & 34.7 & \underline{91.2} & 13.7 \\ \hline
        \rule{0mm}{4mm}3DGraphLLM \cite{zemskova20243dgraphllm} & Objects with 2D, 3D features, Semantic and Spatial Edges & LLAMA3-8B-Instruct & \underline{62.4} & \underline{56.6} & \underline{81.0} & \underline{36.5} & 88.8 & \textbf{15.9 }\\ \hline
        \rule{0mm}{4mm}GPT4Scene-HDM \cite{qi2025gpt4scene} & BEV image with object labels (marks) & Qwen2-VL-7B & \textbf{62.6} & \textbf{57.0} & \textbf{86.3} & \textbf{40.6} & \textbf{96.3} & \underline{15.5} \\ \hline
    \end{tabular}
\end{table}

\begin{table}[!t]
    \caption{Quality of multimodal NeRF-based and Gaussian-Splatting-based map reconstruction on ScanNet dataset \cite{dai2017scannet} for 3D Scene Reconstruction and 3D Semantic Segmentation tasks}
    \label{tab:exp_nerf_splat}
    \centering
    \tiny
    \begin{tabular}{l|l|l|l|l|l|p{2.0cm}}
    \hline
        \rule{0mm}{4mm}Method & Encoded Modalities & PSNR↑ & SSIM↑ & Acc↑ & mIoU↑ & Rendering speed, FPS \\ \hline
        \rule{0mm}{4mm}Semantic Gaussians \cite{guo2024semanticgaussians} & Point cloud with gaussian features, CLIP, LSeg features & 24.67 & 0.69 & 0.67 & 0.49 & \textbf{47.6} \\ \hline
        \rule{0mm}{4mm}uSF \cite{skorokhodov2024usf} & Color, Density, Labels, Uncertainties & 26.81 & 0.85 & 0.98 & 0.90 &  31.3\\ \hline
        \rule{0mm}{4mm}Semantic NeRF \cite{zhi2021semanticnerf} & Color, Density, Labels & 26.74 & 0.80 & 0.98 & 0.89 & - \\ \hline
        \rule{0mm}{4mm}Semantic Ray \cite{liu2023semanticray} & Color, Density, Labels & 26.82 & 0.85 & 0.97 & 0.88  & -\\ \hline
        \rule{0mm}{4mm}NeRFUS \cite{zubkov2025nerfus} & Color, Density, Labels, Uncertainties & \textbf{26.83} & \textbf{0.86} & \textbf{0.99} & \textbf{0.91} &  30.3\\ \hline
    \end{tabular}
\end{table}

\begin{table}[!t]
    \caption{Quality of Encoder-based frame retrieval on Oxford RobotCar dataset \cite{melekhin2024mssplace} for Place recognition task}
    \label{tab:exp_place_recognition}
    \centering
    \tiny
    \begin{tabular}{l|l|l|l}
    \hline
        \rule{0mm}{4mm}Method & Data modalities & AR@1 & AR@1\% \\ \hline
        \rule{0mm}{4mm}NetVLAD \cite{arandjelovic2016netvlad} & Image & 52.54 & 64.62 \\ \hline
        \rule{0mm}{4mm}CosPlace \cite{berton2022cosplace} & Image & 83.46 & 88.86 \\ \hline
        \rule{0mm}{4mm}MixVPR \cite{ali2023mixvpr} & Image & 88.68 & 92.60 \\ \hline
        \rule{0mm}{4mm}EigenPlaces \cite{berton2023eigenplaces} & Image & 83.41 & 88.79 \\ \hline
        \rule{0mm}{4mm}AnyLoc \cite{keetha2023anyloc} & Image & 82.94 & 91.25 \\ \hline
        \rule{0mm}{4mm}PointNetVLAD \cite{uy2018pointnetvlad} & LiDAR Point Cloud & 62.76 & 81.01 \\ \hline
        \rule{0mm}{4mm}LPD-Net \cite{liu2019lpdnet} & LiDAR Point Cloud & 86.28 & 94.92 \\ \hline
        \rule{0mm}{4mm}NDT-Transformer \cite{zhou2021ndttransformer} & LiDAR Point Cloud & 93.80 & 97.65 \\ \hline
        \rule{0mm}{4mm}PPT-Net \cite{hui2021pptnet} & LiDAR Point Cloud & 93.5 & 98.1 \\ \hline
        \rule{0mm}{4mm}TransLoc3D \cite{xu2021transloc3d} & LiDAR Point Cloud & 95.0 & 98.5 \\ \hline
        \rule{0mm}{4mm}ASVT-Net \cite{fan2022svt} & LiDAR Point Cloud & 93.9 & 98.0 \\ \hline
        \rule{0mm}{4mm}MinkLoc3Dv2 \cite{komorowski2022minkloc3dv2} & LiDAR Point Cloud & 96.3 & 98.9 \\ \hline
        \rule{0mm}{4mm}PTC-Net \cite{chen2023ptcnet} & LiDAR Point Cloud & 96.4 & 98.8 \\ \hline
        \rule{0mm}{4mm}BEVPlace++\cite{luo2024bevplace++} & LiDAR Point Cloud & 96.2 & 99.1 \\ \hline
        \rule{0mm}{4mm}AdaFusion \cite{lai2022adafusion} & Image, LiDAR Point Cloud & 98.18 & 99.21 \\ \hline
        \rule{0mm}{4mm}MinkLoc++ \cite{komorowski2021minkloc++} & Image, LiDAR Point Cloud & 97.15 & 99.06 \\ \hline
        \rule{0mm}{4mm}UMF \cite{garcia2024umf} & Image, LiDAR Point Cloud & 98.3 & 99.3 \\ \hline
        \rule{0mm}{4mm}MSSPlace-LI \cite{melekhin2024mssplace} & Image, LiDAR Point Cloud & 98.21 & 99.53 \\ \hline
        \rule{0mm}{4mm}MSSPlace-LIT \cite{melekhin2024mssplace} & Image, LiDAR Point Cloud, Text & 98.22 & 99.53 \\ \hline
        \rule{0mm}{4mm}MSSPlace-LIST \cite{melekhin2024mssplace} & Image, Semantic Label Map, LiDAR Point Cloud, Text & \textbf{98.55} & \textbf{99.64} \\ \hline
    \end{tabular}    
\end{table}

Finally, Table \ref{tab:exp_nerf_splat} demonstrates the promise of using the uncertainty modality for NeRF-based construction of a multimodal 3D map in the NeRFUS \cite{zubkov2025nerfus} approach, since on the one hand it provides a small but significant increase in the quality of semantic scene reconstruction, and on the other hand it allows identifying errors in data markup in the form of masks. The table also shows a significant performance advantage of the Gaussian splatting-based method over NeRF approaches when rendering images from a new angle. Performance measurements were taken on an Nvidia RTX3090 GPU.

\textbf{Multimodal Data Retrieval Module.} 
Table \ref{tab:exp_place_recognition} shows the features of Encoder-based frame retrieval for Place recognition task. It is demonstrated that increasing the number of modalities also gives an increase in quality by Average Recall (AR) metrics for the proposed MSSPlace method \cite{melekhin2024mssplace} on the open Oxford RobotCar  dataset.

The quality of another variant of object retrieval with LLM-based approach BBQ-Deductive \cite{linok2024bbq} is given in Table \ref{tab:exp_llm_retrieval}. It is shown that on open datasets SR3D and NR3D one-stage LLM-based retrieval is significantly inferior to the proposed two-stage LLM-based retrieval, when the first stage selects appropriate objects based on query, and the second stage adds their coordinates to context and runs LLM inference again.

\begin{table}[!ht]
    \caption{Quality of LLM-based object retrieval with BBQ-Deductive \cite{linok2024bbq} approach on SR3D and NR3D datasets}
    \label{tab:exp_llm_retrieval}
    \centering
    \tiny
    \begin{tabular}{l|p{5.0cm}|l|l}
    \hline
        \rule{0mm}{4mm}Method & Scene Reasoning Algorithm & Acc@0.1 Sr3D & Acc@0.1 Nr3D \\ \hline
        \rule{0mm}{4mm}ConceptGraphs \cite{gu2024conceptgraphs} & One-stage LLM-based retrieval from detected objects  & 0.08 & 0.07 \\ \hline
        \rule{0mm}{4mm}ConceptGraphs \cite{gu2024conceptgraphs} & Two-stage LLM-based retrieval: first stage selects appropriate objects based on query, second stage add their coordinates to context & \textbf{0.15} & \textbf{0.12} \\ \hline
    \end{tabular}
\end{table}

\textbf{Application of multimodal 3D maps for state prediction and mobile manipulation.}
Table \ref{tab:exp_predict} shows the effect of using object trajectories (tracklets) (one of the ways to take into account the dynamics in scene graphs) for the task of future map prediction on Waymo Occupancy and Flow Benchmark \cite{ettinger2021waymo}. It is shown that this significantly improves the basic STrajNet method \cite{liu2023strajnet}, which in turn improves the developed OFMPNet method \cite{murhij2024ofmpnet} due to the use of recurrent modules and an original loss function.

\begin{table}[!ht]
    \caption{Quality of future map prediction on Waymo Occupancy and Flow Benchmark~\cite{ettinger2021waymo}}
    \label{tab:exp_predict}
    \centering
    \tiny
    \begin{tabular}{l|p{6.0cm}|l|l|l}
    \hline
        \rule{0mm}{4mm}Method & Trajectory Encoding & Observed AUC ↑  & Occluded AUC ↑ & Flow EPE ↓ \\ \hline
        \rule{0mm}{4mm}STrajNet \cite{liu2023strajnet} & No object trajectory encoding & 0.741 & 0.138 & 3.712 \\ \hline
        \rule{0mm}{4mm}STrajNet \cite{liu2023strajnet} & Trajectory Encoder with Interaction-aware Transformer and Trajectory-aware Cross-attention & 0.751 & 0.161 & \textbf{3.586} \\ \hline
        \rule{0mm}{4mm}OFMPNet \cite{murhij2024ofmpnet} & Trajectory Encoder with Interaction-aware Transformer and Trajectory-aware Cross-attention & \textbf{0.769} & \textbf{0.165} & \textbf{3.586 }\\ \hline
    \end{tabular}
\end{table}

\begin{table}[!ht]
    \caption{Quality of robot action prediction and planning for mobile manipulation}
    \label{tab:exp_mobile_manipulation}
    \centering
    \tiny
    \begin{tabular}{l|l|l}
    \hline
        \rule{0mm}{4mm}Method & Predicted Action  & Success Rate, \% \\ \hline
        \multicolumn{3}{l}{\rule{0mm}{4mm}LAMDEN environment \cite{zhang2025elevnav} (Mobile manipulation in elevator-based multi-floor environment)}\\ \hline
        \rule{0mm}{4mm}LAMDEN (No Scene Graph, Button Detection Only) & Elevator button press & 8.3 \\ \hline
        \rule{0mm}{4mm}LAMDEN (Scene Graph) & Elevator button press & 41.7 \\ \hline
        \rule{0mm}{4mm}LAMDEN (Scene Graph + PCA) & Elevator button press & \textbf{83.3 }\\ \hline
        \multicolumn{3}{l}{\rule{0mm}{4mm}SGN-CIRL environment \cite{oskolkov2025sgncirl} (Indoor Mobile Robot Navigation)} \\ \hline
        \rule{0mm}{4mm}RL-based navigation without scene graph & Navigation to target object & 59.0 \\ \hline
        \rule{0mm}{4mm}SGN-CIRL (target object from scene graph) & Navigation to target object & 55.0 \\ \hline
        \rule{0mm}{4mm}SGN-CIRL (scene graph) & Navigation to target object & \textbf{81.0} \\ \hline
    \end{tabular}
\end{table}

The effect of using the 3D scene graph constructed with the ConceptGraphs-Detect method \cite{gu2024conceptgraphs} on solving the robot action prediction and planning for mobile manipulation problems is shown in Table~\ref{tab:exp_mobile_manipulation}. The table shows that using the scene graph significantly increases the Success rate of executing both the High-level action planning for mobile manipulation in elevator-based multi-floor environment task for the developed LAMDEN method \cite{zhang2025elevnav} and the Low-level indoor mobile robot navigation task for the proposed RL approach SGN-CIRL \cite{oskolkov2025sgncirl}.

\section{Conclusion}  

The article presents the original architecture of the M3DMap modular approach for object-aware multimodal 3D mapping in static and dynamic environments. The implementations of the modules included in this approach are proposed, and the results of experiments confirming their effectiveness for various downstream tasks are presented.

The paper also provides a theoretical justification for the benefits of using various modalities in the Object segmentation and tracking module, as well as the use of additional data processing in it for feeding along with raw sensory data to the Module for 3D map construction and updating.

The limitations of the proposed M3DMap approach include the need to synchronize all modules, as well as difficulties in ensuring its operation in real time.

Future work should include the study of both individual modules and the entire method on various benchmarks containing dynamic scenes and detailed markup to check the quality of all downstream tasks at once. A special role is seen in the creation of photorealistic simulation environments for automated dataset generation.

\bibliographystyle{plainnat}
\bibliography{paper}

\end{document}